\documentclass{article}

\usepackage{iclr2027_conference,times}

\usepackage[utf8]{inputenc}
\usepackage[T1]{fontenc}
\usepackage{hyperref}
\hypersetup{hidelinks}
\usepackage{url}
\usepackage{booktabs}
\usepackage{amsfonts}
\usepackage{amsmath}
\usepackage{amssymb}
\usepackage{nicefrac}
\usepackage{microtype}
\usepackage{xcolor}
\usepackage{graphicx}
\usepackage{subcaption}
\usepackage{multirow}
\usepackage{wrapfig}
\usepackage{float}
\usepackage{placeins}
\usepackage{pifont}

\usepackage[ruled]{algorithm2e}
\usepackage{comment}
\newcounter{algstep}

\usepackage{makecell}
\title{ZeroCode: On-demand Error-Correcting Code Construction from the Zero Matrix via Reinforcement Learning}

\author{
Ju-Hyeong Lee\\
University of Ulsan\\
\texttt{xkca2446@gmail.com}
\And
Yongjune Kim\\
Pohang University of Science and Technology (POSTECH)\\
\texttt{yongjune@postech.ac.kr}
\And
Sang-Hyo Kim\\
Sungkyunkwan University\\
\texttt{iamshkim@skku.edu}
\AND
Dae-Young Yun\\
University of Ulsan\\
\texttt{yundy@mail.ulsan.ac.kr}
\And
Hee-Youl Kwak\\
University of Ulsan\\
\texttt{ghy1228@gmail.com}
}

\iclrfinalcopy

\begin{document}

\maketitle

\begin{abstract}
Error-correcting codes (ECCs) are essential across diverse applications—from wireless communications and storage to quantum computing—yet each application imposes distinct design requirements on the parity-check matrix (PCM). To address these on-demand requirements in a unified framework, we propose {\em ZeroCode}\footnote{Code is available at \url{https://github.com/wngud387/ppo_code}.}, a reinforcement learning (RL)-based approach that constructs PCMs sequentially from the all-zero matrix. ZeroCode formulates construction as a discrete sequential decision-making problem and uses proximal policy optimization with action masking to select valid edges. ZeroCode achieves a gain of approximately 1 dB over the prior RL-based construction method at a bit error rate (BER) of $10^{-4}$ for the (32,16) code and outperforms existing genetic, differentiable, and classical code-design methods in our experiments. Beyond optimizing decoding performance, the masking mechanism allows on-demand structural constraints—such as a maximum degree, 4-cycle-free structure, and quasi-cyclic structure—to be flexibly incorporated. Moreover, a single policy rollout yields a library of PCMs with varying edge counts, offering trade-offs between decoding performance and complexity without retraining. Overall, ZeroCode addresses diverse code-design requirements within a unified framework, providing solutions with optimized decoding performance under given constraints.

\end{abstract}

\section{Introduction}
Error-correcting codes (ECCs) are a foundational technology for ensuring reliability, and their applications continue to expand across diverse domains including communication systems~\citep{richardson2018design}, memory devices~\citep{zhao2013ldpc}, distributed storage~\citep{ramkumar2022codes}, and quantum computing~\citep{shor1995scheme}. While decoding performance is universally important, each application imposes distinct design requirements on the parity-check matrix (PCM), such as complexity constraints, efficient encodability, and rate compatibility. No single hand-crafted code family can simultaneously satisfy this diverse set of requirements, motivating a flexible, on-demand approach to PCM design.

Early design for low-density parity-check (LDPC) codes primarily focused on asymptotic performance~\citep{gallager1962low, mackay1999good, richardson2001design}: Degree distributions were optimized as proxies for PCMs to avoid a combinatorial search over individual PCMs while maximizing performance in the infinite-blocklength regime. While effective for long block lengths, this approach does not directly optimize the metric most relevant in practice—the bit error rate (BER) of a finite-length code. 
More recent work has addressed finite-length code design by optimizing the PCM itself, using genetic algorithms (GA)~\citep{elkelesh2019decoder}, differentiable belief propagation (BP)~\citep{choukroun2024factor}, or reinforcement learning (RL)~\citep{tian2025gnn}, with error performance as the optimization objective.
Such finite-length optimization is particularly well-suited to short block lengths for low-latency communications~\citep{yue2023efficient, shirvanimoghaddam2018short}, where asymptotic metrics are less informative.

Among these approaches, the RL-based approach of~\citet{tian2025gnn} is most closely related to our work, yet it has several fundamental limitations. It employs deep deterministic policy gradient (DDPG)~\citep{lillicrap2015continuous}, an algorithm designed for continuous action spaces, whereas PCM construction is inherently discrete—each edge is either included or excluded. This mismatch can result in inaccurate gradient estimates~\citep{chen2019large}. Furthermore, the method initializes from a non-zero matrix and updates multiple edges at each step, making optimization sensitive to initialization and the search trajectory difficult to control.

In this paper, we address these limitations with {\em ZeroCode}, an RL framework that constructs PCMs sequentially by adding one edge per step from the all-zero matrix. This formulation removes dependence on the choice of a pre-constructed initial PCM and enables stable learning. It also naturally induces a Markov decision process (MDP) formulation that facilitates the incorporation of design-specific constraints via masking. We employ proximal policy optimization (PPO) to learn the policy for this discrete sequential construction problem. Experimental results show that ZeroCode significantly outperforms the existing RL-based construction~\citep{tian2025gnn}, establishing a new state of the art among RL-based approaches to code design. Compared with non-RL approaches~\citep{richardson2001design, elkelesh2019decoder, choukroun2024factor}, ZeroCode also achieves superior BER performance at short block lengths.

Beyond optimizing performance, ZeroCode incorporates \emph{on-demand} structural constraints directly into the construction process via \emph{action masking}. Moreover, a single rollout of the trained policy generates a library of codes spanning a wide range of edge counts, yielding multiple performance--complexity trade-off points without retraining. In contrast, prior approaches~\citep{elkelesh2019decoder, choukroun2024factor} focus on optimizing a target PCM rather than explicitly constructing a library indexed by edge count.

The main contributions of this paper are as follows.
\begin{enumerate}
\item We formulate PCM construction as a sequential decision-making problem over a discrete action space and learn an edge-by-edge construction policy starting from the all-zero matrix.
\item ZeroCode significantly outperforms the existing RL-based method and achieves superior BER performance over non-RL approaches at block lengths $n \in \{64, 128, 256\}$.
\vspace{-1pt}
\item Structural constraints such as 4-cycle-free graphs, quasi-cyclic structure, maximum degree, and rate compatibility can be flexibly incorporated via masking, enabling on-demand code design across diverse application requirements.
\vspace{-1pt}
\item A single trained policy yields a library of PCMs spanning a wide range of edge counts along its rollout trajectory, providing favorable performance--complexity trade-offs across operating points.
\end{enumerate}

\section{Related work}
\citet{richardson2001design} optimized degree distributions for asymptotic performance, while the progressive edge-growth (PEG) algorithm~\citep{hu2005regular} constructs finite-length Tanner graphs. More recent methods optimize finite-length codes for a specified decoder: \citet{elkelesh2019decoder} used block error rate as the fitness function of a GA, while \citet{choukroun2024factor} optimized the PCM through differentiable BP. However, those works do not explicitly address the range of on-demand structural constraints considered in ZeroCode.
The RL-based approach of~\citet{tian2025gnn} formulates PCM construction as an MDP and learns a DDPG-based policy that modifies multiple edges per step from a pre-constructed PCM. ZeroCode differs in three respects: (i) it constructs PCMs edge by edge from the all-zero matrix; (ii) it employs PPO, which is better suited to the discrete action space; and (iii) it uses action masking to enforce structural constraints. \citet{huang2020ai} also applied RL and GA to the construction of polar codes and other linear block codes, but did not consider enforcing PCM-specific structural constraints during the construction process.

A complementary line of research applies deep learning and RL to improve the decoding algorithm, including neural BP decoders~\citep{nachmani2018deep, kwak2023boosting, kwak2025boosted}, transformer-based decoders~\citep{choukroun2022error, choukroun2024foundation, park2025multiple, park2024crossmpt}, and RL-based decoding schedules~\citep{habib2020learning, habib2023reldec}. These methods are orthogonal to ours and can be combined with ZeroCode.

Beyond ECC, RL has proven effective in complex sequential decision-making over discrete spaces, with notable successes including game playing~\citep{silver2016mastering, berner2019dota} and chip placement~\citep{mirhoseini2021graph}. The closest analogy is AlphaChip~\citep{mirhoseini2021graph}, which constructs chip layouts incrementally from an initially empty canvas, demonstrating the effectiveness of sequential construction for stable optimization and constraint-aware design. ZeroCode adopts the same incremental construction paradigm for PCM design.

\begin{figure}[t]
  \centering
  \begin{subfigure}[b]{0.96\linewidth}
    \centering
    \includegraphics[width=\linewidth]{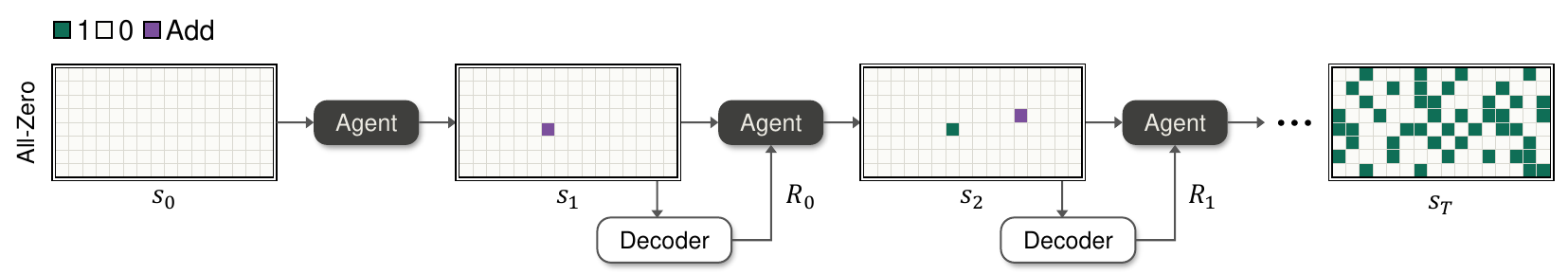}
    \caption{}
    \label{fig:overview_zerocode}
  \end{subfigure}

  \vspace{0.25em}

  \begin{subfigure}[b]{0.96\linewidth}
    \centering
    \includegraphics[width=\linewidth]{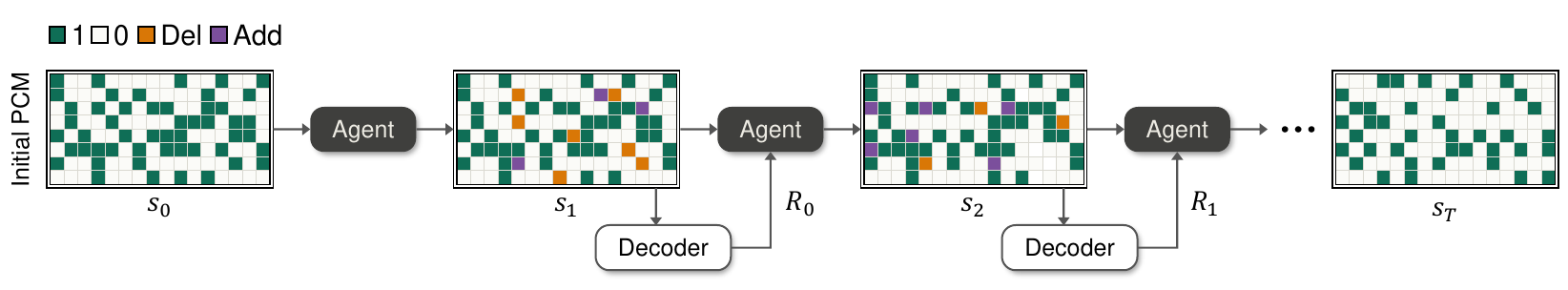}
    \caption{}
    \label{fig:overview_ddpg}
  \end{subfigure}
  \caption{Overview of (a) ZeroCode compared with (b) the prior RL-based
  approach, DDPG~\citep{tian2025gnn}. DDPG starts from a pre-constructed PCM
  and updates multiple edges at each step, whereas ZeroCode begins with an
  all-zero matrix and the agent incrementally adds a single edge per step. In
  both methods, the PCM at each step serves as the state, and the BER obtained
  from the decoder is used as the reward signal.}
  \label{fig:construction_comparison}
\end{figure}

\section{Background}

\subsection{Error-correcting codes and parity-check matrix}

An $(n,k)$ code encodes an information message of length $k$ into a codeword of length $n$ to improve reliability against channel noise. A PCM $\mathbf{H} \in \{0,1\}^{m \times n}$ defines the code, where each row represents a parity-check constraint and each column corresponds to a coded bit. The set of valid codewords is the null space of $\mathbf{H}$ over $\mathrm{GF}(2)$. When $\mathbf{H}$ has full rank $m$, the information length is $k = n - m$ and the code rate is $r = k/n$. The goal of this work is to construct PCMs that yield strong decoding performance while satisfying application-specific structural constraints.

\subsection{Decoding and performance evaluation}

Given a PCM $\mathbf{H}$, a decoder estimates the transmitted codeword from the received noisy channel output, and its performance serves as the reward signal for our RL framework. While any decoder can in principle be used, we adopt BP as a representative example. BP iteratively exchanges LLR messages over the Tanner graph of $\mathbf{H}$ consisting of $n$ variable nodes (VNs) and $m$ check nodes (CNs). The message updates at the $\ell$-th iteration are given by
\begin{equation}
{\hat m}^{(\ell)}_{c\rightarrow v}=2\tanh^{-1}\!\left(\prod_{v'\in\mathcal{N}(c)\setminus\{v\}}\tanh\!\left(\frac{m^{(\ell-1)}_{v'\rightarrow c}}{2}\right)\right),\quad
m^{(\ell)}_{v\rightarrow c}=m_v^{\rm ch}+\sum_{c'\in\mathcal{N}(v)\setminus\{c\}}{\hat m}^{(\ell)}_{c'\rightarrow v},
\label{eq:bg_bp_updates}
\end{equation}
where $\mathcal{N}(x)$ denotes the set of neighbor nodes of node $x$ and $m_{v}^{\rm ch}$ is the channel LLR at VN $v$. At each iteration, the output LLR at VN $v$ is
${\bar m}_{v}^{(\ell)} = m_{v}^{\rm ch} + \sum_{c \in \mathcal{N}(v)} {\hat m}^{(\ell)}_{c\rightarrow v}$,
and the hard decision is $0$ if ${\bar m}_{v}^{(\ell)} > 0$ and $1$ otherwise. Decoding terminates when all parity checks are satisfied or the maximum number of iterations is reached. We evaluate each PCM in terms of the BER by the BP decoder.

\subsection{Proximal policy optimization with action masking}
\label{sec:masked_ppo}
RL models sequential decision-making as an MDP defined over a state space $\mathcal{S}$, an action space $\mathcal{A}$, and a reward function $R$.
At each time step $t$, the agent observes a state $s_t \in \mathcal{S}$, selects an action $a_t \in \mathcal{A}$ according to its policy $\pi_\theta(\cdot \mid s_t)$ parameterized by $\theta$, transitions to the next state $s_{t+1}$, and receives a reward $R_t$.
The goal of RL is to learn a policy that maximizes the expected cumulative return.

We adopt PPO~\citep{schulman2017proximal} to learn the PCM construction policy. 
PPO is an actor--critic algorithm that jointly learns a policy $\pi_\theta(\cdot \mid s_t)$ and a value function $V_\phi(s_t)$. 
At each time step, an action is sampled as $a_t \sim \pi_\theta(\cdot \mid s_t)$, and the resulting transition $\tau_t := (s_t, a_t, R_t, s_{t+1})$ is stored in a rollout buffer $\mathcal{D}$. 
PPO updates $(\theta, \phi)$ using samples from $\mathcal{D}$ by minimizing a combined loss $L^{\mathrm{PPO}}(\theta, \phi)$ consisting of a clipped policy loss, a value loss, and an entropy bonus. 
We refer to~\citet{schulman2017proximal} for the precise definition of each term.

In our PCM construction problem, the valid action set varies with the current state: previously selected edges cannot be selected again, and edges that violate application-specific structural constraints are also excluded. 
We therefore combine PPO with action masking so that actions disallowed in the current state are never sampled. 
Let $\mathcal{A}_{\mathrm{valid}}(s_t) \subseteq \mathcal{A}$ denote the valid action set at state $s_t$. 
The masked policy $\tilde{\pi}_\theta(a \mid s_t)$ is obtained by renormalizing $\pi_\theta(a \mid s_t)$ over $\mathcal{A}_{\mathrm{valid}}(s_t)$:
\begin{equation}
\tilde{\pi}_\theta(a \mid s_t) =
\frac{
\pi_\theta(a \mid s_t)\,\mathbf{1}[a \in \mathcal{A}_{\mathrm{valid}}(s_t)]
}{
\sum_{a' \in \mathcal{A}}
\pi_\theta(a' \mid s_t)\,\mathbf{1}[a' \in \mathcal{A}_{\mathrm{valid}}(s_t)]
}.
\label{eq:masked_policy}
\end{equation}
As shown by~\citet{huang2020closer}, when the mask is independent of the policy parameter $\theta$, the resulting policy-gradient estimator is unbiased for the gradient of the masked policy. 
Therefore, PPO can be applied with $\tilde{\pi}_\theta$ in place of $\pi_\theta$ without modifying the objective. 
This compatibility allows ZeroCode to incorporate structural constraints directly through masking while retaining the standard PPO training procedure.

\begin{figure}[t]
  \centering
  \includegraphics[width=1.\linewidth]{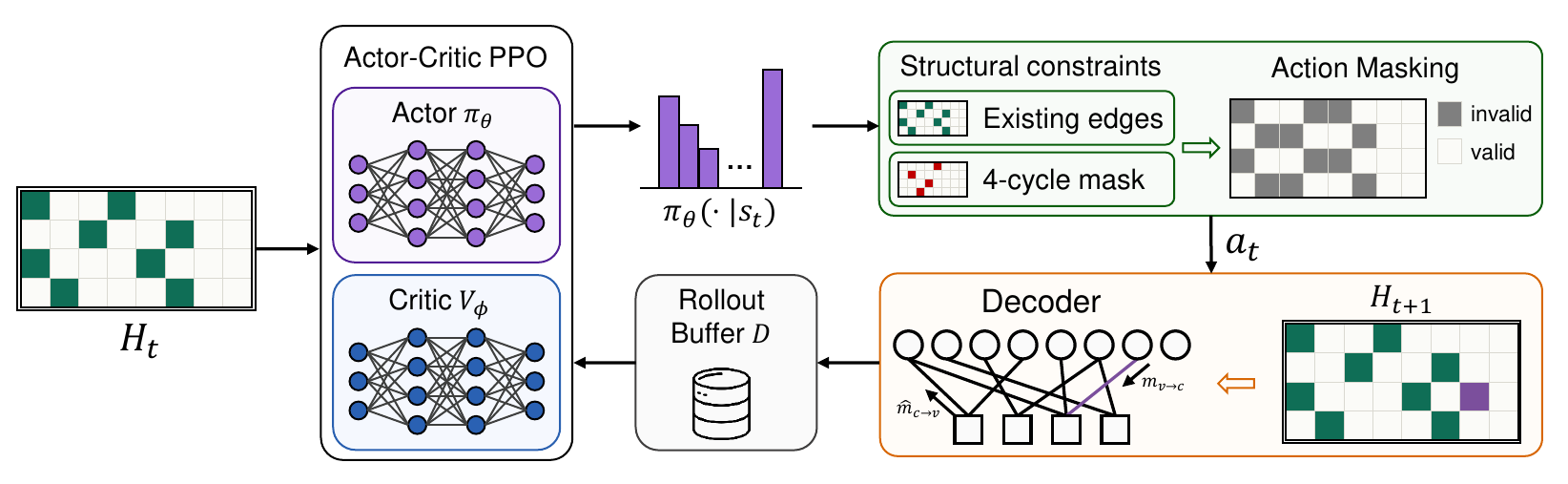}
  \caption{Overview of the PCM construction process in ZeroCode. The agent receives the current PCM as the state, applies an action mask to retain only edge candidates that satisfy the structural constraints, and samples a single edge from the masked policy. The updated matrix is evaluated by the BP decoder to produce the reward signal. Starting from the all-zero matrix, ZeroCode repeats this process to optimize PCMs.}
  \label{fig:framework}
\end{figure}

\section{Proposed method}
\label{sec:method}
We formulate the construction of a binary PCM $\mathbf{H} \in \{0,1\}^{m \times n}$ as an RL problem. Starting from the all-zero matrix, the agent incrementally constructs $\mathbf{H}$ by selecting one edge at a time and setting it to $1$, with the objective of optimizing decoding performance throughout the construction process. We define a PCM as valid when it has full rank over $\mathrm{GF}(2)$ and contains neither an all-zero row nor an all-zero column.

\subsection{MDP definition for ZeroCode}
\label{sec:mdp}
In this subsection, we define the state, action, and reward for PCM construction. An episode begins from the all-zero matrix and consists of a sequence of edge additions performed by the agent until the termination condition is satisfied. Each individual edge addition within an episode corresponds to one step. The state $s_t$ at step $t$ is the current binary PCM, $s_t = \mathbf{H}_t \in \{0,1\}^{m \times n}$, with the initial state set to the all-zero matrix $\mathbf{H}_0 = \mathbf{0}_{m \times n}$. At each step, the agent selects a position in the matrix and adds an edge there. The episode terminates when no valid edge remains in the masked action set.

Every valid PCM encountered along the construction trajectory is retained in the code library. Because one edge is added at each step, the retained PCMs span different edge counts and, consequently, different decoding complexities. The resulting library allows an appropriate PCM to be selected according to application requirements.

After each step, BP decoding is run on the updated matrix $\mathbf{H}_{t+1}$ to compute $\mathrm{BER}_{t+1}$. We define $\Lambda(\mathbf{H}_{t+1}) = -\ln \mathrm{BER}_{t+1}$ and use it as the reward:
\begin{equation}
R_t = R(s_t, a_t)
= \Lambda(\mathbf{H}_{t+1}).
\label{eq:reward}
\end{equation}

\subsection{Action masking}
\label{sec:masking}

In ZeroCode, action masking is used to restrict the valid action set at each step. We construct an action mask matrix $\mathbf{M}(\mathbf{H}_t) \in \{0,1\}^{m \times n}$, where $\mathbf{M}(\mathbf{H}_t)_{i,j} = 1$ indicates that edge $(i,j)$ cannot be selected. By default, $\mathbf{M}(\mathbf{H}_t) = \mathbf{H}_t$, since edges, once added, cannot be removed. Additional positions can be masked to enforce structural constraints. The valid action set is then given by
\[
\mathcal{A}_{\mathrm{valid}}(s_t)
=
\{(i,j) : \mathbf{M}(\mathbf{H}_t)_{i,j} = 0\}.
\]
Beyond enforcing application-specific structural constraints, action masking also reduces the action space by pruning structurally invalid or undesirable candidates a priori, thereby facilitating more efficient policy learning.

\paragraph{4-cycle masking.}
It is well known that BP decoding performance is degraded by short cycles 
in the Tanner graph, particularly 4-cycles. Short cycles undermine the 
independence of the extrinsic information exchanged during iterative 
decoding, which in turn degrades decoding 
performance~\citep{ryan2009channel, fan2008design}. Accordingly, at 
each step we identify positions $(i, j)$ that would introduce a new 
4-cycle in the Tanner graph of $\mathbf{H}_t$, and mask all such positions from $\mathcal{A}_{\mathrm{valid}}(s_t)$. 
Specifically, selecting position $(i, j)$ creates a new 4-cycle 
if there exists a row $i' \neq i$ and a column $j' \neq j$ such that 
$(\mathbf{H}_t)_{i', j} = (\mathbf{H}_t)_{i, j'} = (\mathbf{H}_t)_{i', j'} = 1$.

For example, consider the following PCM $\mathbf{H}_t$ and its corresponding mask matrix $\mathbf{M}(\mathbf{H}_t)$:
\begin{equation}
\mathbf{H}_t=
\begin{bmatrix}
1 & 0 & 0 & 1 & 0 & 0 & 0 & 0\\
0 & 0 & 1 & 0 & 0 & 1 & 0 & 0\\
1 & 0 & 0 & 0 & 1 & 0 & 0 & 0\\
0 & 1 & 0 & 0 & 0 & 1 & 0 & 0
\end{bmatrix},
\quad
\mathbf{M}(\mathbf{H}_t)=\mathbf{H}_t \vee
\begin{bmatrix}
0 & 0 & 0 & 0 & 1 & 0 & 0 & 0\\
0 & 1 & 0 & 0 & 0 & 0 & 0 & 0\\
0 & 0 & 0 & 1 & 0 & 0 & 0 & 0\\
0 & 0 & 1 & 0 & 0 & 0 & 0 & 0
\end{bmatrix},
\label{eq:4cycle_example}
\end{equation}
where $\vee$ denotes the element-wise OR operation, the first term masks existing edges, and the second term masks positions that would introduce new 4-cycles. This masking prevents exploration of PCMs containing 4-cycles throughout training.

\paragraph{QC masking.}
To enforce a quasi-cyclic (QC) structure, we partition the PCM $\mathbf{H} \in \{0,1\}^{m \times n}$ into blocks of size $z \times z$, where $z$ denotes the lifting size. When the agent selects an individual edge, all other edges with the same cyclic-shift relation within the corresponding block are automatically added to form a circulant permutation matrix. Edge selections that would violate this block structure are masked out from $\mathcal{A}_{\mathrm{valid}}(s_t)$, thereby preserving the QC structure of the PCM throughout the construction process.

\paragraph{Max-degree masking.}
The maximum variable-node degree directly determines the hardware complexity of the BP decoder, so applications often require PCMs of the same size under different degree constraints \citep{richardson2001design}. To incorporate this requirement, for a prescribed maximum variable-node degree $d_{\max}$, at each step we identify any variable node $v$ whose current degree has reached $d_{\max}$ and mask the entire corresponding column from $\mathcal{A}_{\mathrm{valid}}(s_t)$. This ensures that every generated PCM satisfies the degree constraint by construction. Moreover, max-degree masking can also be applied at inference time without retraining, as discussed in Section~\ref{sec:on_demand}.

\subsection{Policy optimization in ZeroCode}
\label{sec:policy_optimization}
ZeroCode implements the policy network $\pi_\theta$ and value network $V_\phi$ as two independent MLPs. Both networks take the flattened PCM $\mathrm{vec}(\mathbf{H}_t) \in \{0,1\}^{mn}$ as input. Let $f_\theta$ and $g_\phi$ denote the MLPs that produce the policy logits and the value estimate, respectively. The unmasked policy and the state-value estimate are then given by
\begin{equation}
\pi_\theta(\cdot \mid s_t)
=
\mathrm{softmax}\!\left(f_\theta(\mathrm{vec}(\mathbf{H}_t))\right),
\qquad
V_\phi(s_t)
=
g_\phi(\mathrm{vec}(\mathbf{H}_t)).
\label{eq:actor_critic_mlp}
\end{equation}

As shown in the architectural ablation study in Appendix~\ref{sec:architecture_ablation}, the MLP-based actor--critic achieves more stable training and better decoding performance than GNN-based variants. We therefore adopt this MLP architecture in ZeroCode.

\noindent
\begin{minipage}[t]{0.45\textwidth}
\vspace{0pt}

The masked policy $\tilde{\pi}_{\theta}(\cdot \mid s_t)$ is obtained~from $\pi_{\theta}(\cdot\mid s_t)$ by assigning zero probability~to all masked actions and renormalizing over $\mathcal{A}_{\mathrm{valid}}(s_t)$, as defined in Eq.~\eqref{eq:masked_policy}. Algorithm~\ref{alg:ppo} summarizes the training procedure of ZeroCode. Each episode starts from the all-zero matrix $\mathbf{H}_0 = \mathbf{0}_{m \times n}$, and at every step, ZeroCode samples one valid edge from the masked policy $\tilde{\pi}_\theta$ and adds it to the current PCM. The updated PCM is evaluated by BP decoding, and the resulting transition is stored in the rollout buffer. When an episode terminates, the next episode is initialized from the all-zero matrix. Once the rollout buffer is filled, the policy and value networks are updated by minimizing the PPO loss over the collected transitions. The detailed pseudocode and hyperparameters are provided in Appendix~\ref{app:training_details}.
\end{minipage}\hfill
\begin{minipage}[t]{0.525\textwidth}
\vspace*{-1.75em}
\begin{algorithm}[H]
\caption{Training procedure of \mbox{ZeroCode}}
\label{alg:ppo}
\DontPrintSemicolon
\textbf{Input:} PCM size $(m,n)$, structural constraints\\
\textbf{Initialize:} $\pi_\theta$, $V_\phi$, and $\mathcal{D}\leftarrow\emptyset$
\setcounter{algstep}{0}
\begin{list}{\arabic{algstep}:}{%
  \usecounter{algstep}%
  \setlength{\leftmargin}{1.5em}%
  \setlength{\rightmargin}{0pt}%
  \setlength{\itemindent}{0pt}%
  \setlength{\labelwidth}{1.5em}%
  \setlength{\labelsep}{0.3em}%
  \setlength{\itemsep}{3pt}%
  \setlength{\parsep}{0pt}%
  \setlength{\topsep}{0pt}%
}
\item \textbf{Start:} Initialize each episode with $\mathbf{H}_0=\mathbf{0}_{m\times n}$.

\item \textbf{Construction:} At step $t$, construct the action mask $\mathbf{M}(\mathbf{H}_t)$ to reflect the structural constraints. Sample $a_t \sim \tilde{\pi}_\theta(\cdot \mid \mathbf{H}_t)$ from the resulting masked policy, and add the selected edge to obtain $\mathbf{H}_{t+1}$.

\item \textbf{Evaluation:} Evaluate $\mathbf{H}_{t+1}$ by BP decoding, retain it in the code library if it is valid, assign $R_t$ according to Eq.~\eqref{eq:reward}, and store the transition in $\mathcal{D}$. If the episode terminates, return to the Start step; otherwise, proceed to the next step.

\item \textbf{Update:} Once the rollout buffer $\mathcal{D}$ is full, update $\pi_\theta$ and $V_\phi$ by minimizing $L^{\mathrm{PPO}}$ over $\mathcal{D}$, set $\mathcal{D}\leftarrow\emptyset$, and resume the episode from where it was paused.
\end{list}
\textbf{Output:} $\pi_\theta$
\end{algorithm}
\end{minipage}
\par\vspace{0.5em}

\begin{figure}[t]
  \centering

  \begin{subfigure}[b]{0.37\textwidth}
    \centering
    \includegraphics[width=\linewidth]{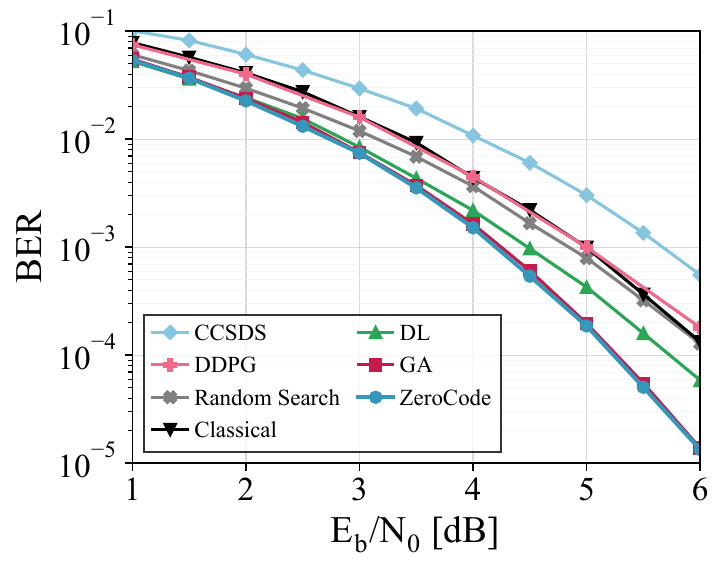}
    \caption{}
    \label{fig:plot_32}
  \end{subfigure}
  \hspace{0.03\textwidth}
  \begin{subfigure}[b]{0.37\textwidth}
    \centering
    \includegraphics[width=\linewidth]{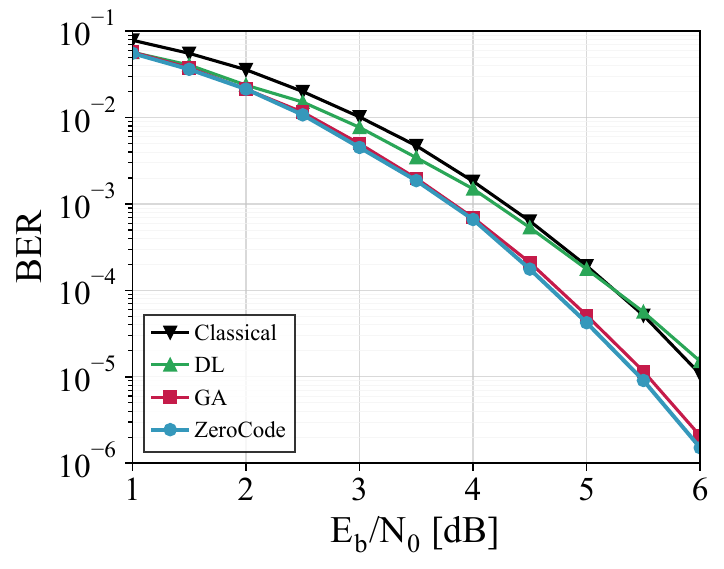}
    \caption{}
    \label{fig:plot_64}
  \end{subfigure}

  \vspace{0.1em}

  \begin{subfigure}[b]{0.37\textwidth}
    \centering
    \includegraphics[width=\linewidth]{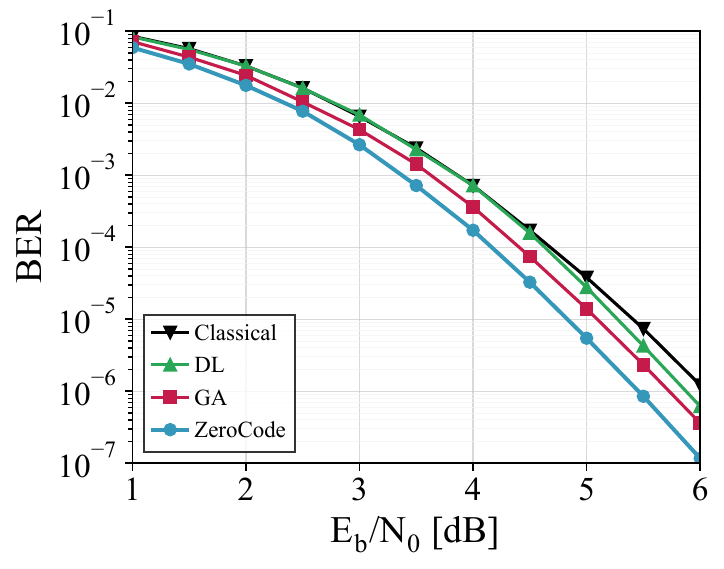}
    \caption{}
    \label{fig:plot_128}
  \end{subfigure}
  \hspace{0.03\textwidth}
  \begin{subfigure}[b]{0.37\textwidth}
    \centering
    \includegraphics[width=\linewidth]{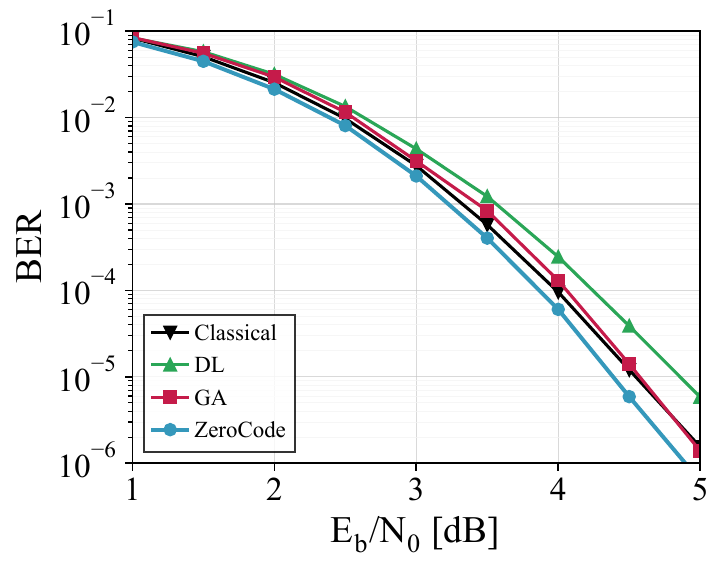}
    \caption{}
    \label{fig:plot_256}
  \end{subfigure}

  \caption{Comparison of the codes constructed by ZeroCode with DDPG~\citep{tian2025gnn}, GA~\citep{elkelesh2019decoder}, DL~\citep{choukroun2024factor}, and the classical construction~\citep{richardson2001design} for various code dimensions over the AWGN channel: (a)~$(32,16)$, (b)~$(64,32)$, (c)~$(128,64)$, and (d)~$(256,128)$.}
  \label{fig:performance}
\end{figure}

\section{Experiments}
\label{sec:experiments}
\subsection{Experimental setup}
We consider block lengths $n \in \{32, 64, 128, 256\}$ and a code rate of $r = 0.5$ over the additive white Gaussian noise (AWGN) channel. The maximum number of BP iterations is set to $8$ for $n = 32$ and $5$ for all other block lengths. During training, the reward is computed based on the BER measured at ${\rm E}_{\rm b}/{\rm N}_0 = 5.0$~dB.

\subsection{Performance comparison}
Figure~\ref{fig:performance} illustrates the BER performance of ZeroCode across four block lengths. For the $(32,16)$ code in Figure~\ref{fig:plot_32}, ZeroCode is compared against DDPG~\citep{tian2025gnn}, GA~\citep{elkelesh2019decoder}, deep learning (DL)~\citep{choukroun2024factor}, classical~\citep{richardson2001design}, Random Search, and the CCSDS code~\citep{ccsds2015short}. The classical baseline is constructed from the degree distribution of~\citet{richardson2001design} using the PEG algorithm~\citep{hu2005regular}. The Random Search baseline starts from the all-zero matrix and adds edges at random positions until a valid PCM is obtained, providing a learning-free reference.

Among these, DDPG~\citep{tian2025gnn} is the most direct point of comparison, as it shares with ZeroCode the RL-based formulation for PCM construction. In contrast, GA and DL represent independent lines of work based on evolutionary search and differentiable BP, respectively, and serve as reference points for data-driven PCM design. Compared with DDPG, ZeroCode yields a clear improvement of approximately $1$~dB at $\mathrm{BER}=10^{-4}$ for the $(32,16)$ code, demonstrating the effectiveness of the proposed sequential construction formulation. ZeroCode also outperforms Random Search, suggesting that the learned policy contributes to the observed performance gains.

Compared with DL~\citep{choukroun2024factor}, ZeroCode achieves substantially better performance across all configurations. Against GA~\citep{elkelesh2019decoder}, ZeroCode achieves comparable performance at $n=32$ and increasingly outperforms it as the block length grows. Moreover, ZeroCode can directly incorporate diverse structural constraints through masking, as demonstrated in later subsections.

\subsection{Component Ablation}
\label{sec:component_ablation}

\begin{table}[H]
\centering
\caption{Ablation study of ZeroCode on the $(32,16)$ code.}
\label{tab:mdp_ablation}
\setlength{\tabcolsep}{5pt}
\begin{tabular}{lcccc}
\toprule
Method
& RL Algorithm
& Initial PCM
& Edge Update
& BER (${\rm E}_{\rm b}/{\rm N}_0 = 5.0$~dB) \\
\midrule
(A) \citep{tian2025gnn}
& DDPG
& CCSDS
& Multi-Flip
& $1.0 \times 10^{-3}$ \\

(B)
& PPO
& CCSDS
& Multi-Flip
& $2.1 \times 10^{-3}$ \\

(C)
& PPO
& CCSDS
& Single-Flip
& $1.0 \times 10^{-3}$ \\

\textbf{ZeroCode}
& \textbf{PPO}
& \textbf{All-zero}
& \textbf{Single-Add}
& $\mathbf{2.2 \times 10^{-4}}$ \\

\bottomrule
\end{tabular}
\end{table}

To examine the contribution of ZeroCode's components, Table~\ref{tab:mdp_ablation} compares variants with different RL algorithms, initial PCMs, and edge update schemes. First, comparing (A) and (B) shows that replacing DDPG with PPO alone does not improve performance: with the other components unchanged, the BER increases from $1.0\times10^{-3}$ to $2.1\times10^{-3}$. Thus, the performance gain of ZeroCode cannot be attributed simply to the choice of PPO.

Next, replacing Multi-Flip in (B) with Single-Flip in (C) reduces the BER from $2.1\times10^{-3}$ to $1.0\times10^{-3}$. This improvement is consistent with the finer-grained credit assignment enabled by single-edge updates: each reward reflects one edge modification rather than the combined effect of multiple modifications, potentially making individual action effects easier to learn.

Finally, comparing (C) with ZeroCode reveals the largest gain among these comparisons. Variant (C) modifies a pre-constructed CCSDS PCM through single-edge flips, whereas ZeroCode constructs a PCM from the all-zero matrix through single-edge additions. This change reduces the BER from $1.0\times10^{-3}$ to $2.2\times10^{-4}$. These results suggest that ZeroCode's performance improvement over the prior RL-based method primarily stems from reformulating the problem from modifying a pre-constructed PCM to sequentially constructing a PCM from the all-zero matrix.

\begin{figure}[t]
  \centering
  \includegraphics[width=1.\linewidth]{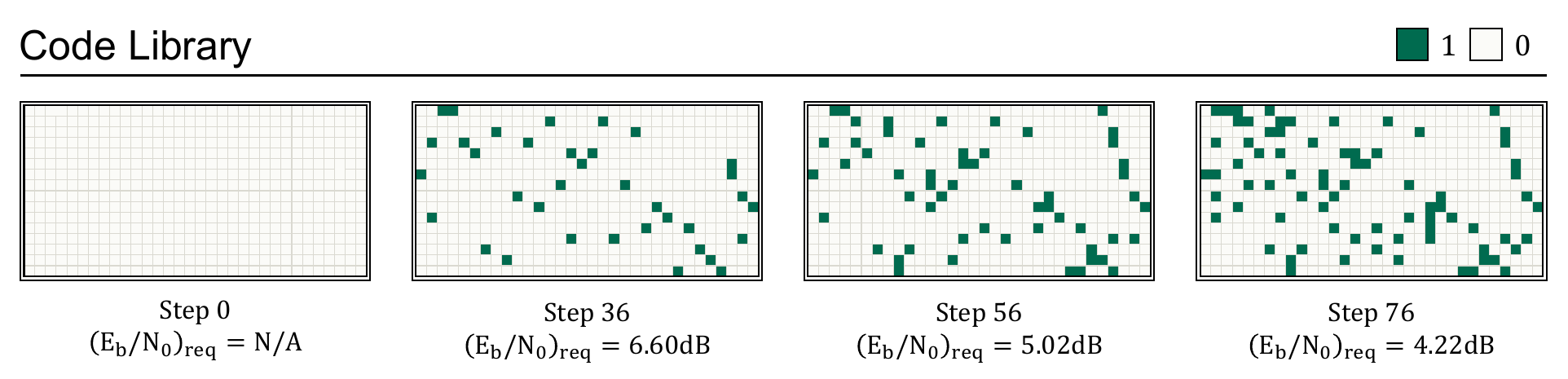}
  \caption{Sequential construction of a code library using a single trained ZeroCode policy for the $(32,16)$ code. The first matrix shows the all-zero initialization, while the remaining matrices show valid PCMs obtained along the construction trajectory, together with their step indices and the required $({\rm E}_{\rm b}/{\rm N}_0)_{\rm req}$ for achieving $\mathrm{BER}=10^{-3}$.}
  \label{fig:matrix_construction}
\end{figure}

\subsection{A library of codes from a single rollout}
\label{sec:library}

In this subsection, we show that a single rollout of a trained policy can generate multiple PCMs with different edge counts, forming a code library. Figure~\ref{fig:matrix_construction} visualizes a single rollout of the trained policy for the $(32,16)$ code. Starting from the all-zero matrix, ZeroCode adds one edge at each step, so the edge count and decoding complexity gradually increase as construction proceeds. At the same time, the required ${\rm E}_{\rm b}/{\rm N}_0$ to achieve $\mathrm{BER}=10^{-3}$, denoted by $({\rm E}_{\rm b}/{\rm N}_0)_{\rm req}$, generally decreases. The valid PCMs generated along this construction trajectory form a code library with different edge counts.

Figure~\ref{fig:library} shows the relationship between the edge count and $({\rm E}_{\rm b}/{\rm N}_0)_{\rm req}$ for the resulting code library. As the edge count increases, decoding complexity increases while a lower $({\rm E}_{\rm b}/{\rm N}_0)_{\rm req}$ can be achieved, yielding a performance--complexity trade-off curve from a single rollout. Therefore, given an edge-count constraint imposed by an application, an appropriate PCM can be selected from the library without retraining. Moreover, the ZeroCode trade-off curve in Figure~\ref{fig:library} lies below the operating points of the existing methods, demonstrating that ZeroCode spans a wide range of edge counts while achieving competitive decoding performance across operating points.


\subsection{Maximum degree constraint}
\label{sec:on_demand}
\begin{figure}[t]
  \centering
  \begin{subfigure}[b]{0.33\textwidth}
    \centering
    \includegraphics[width=\linewidth]{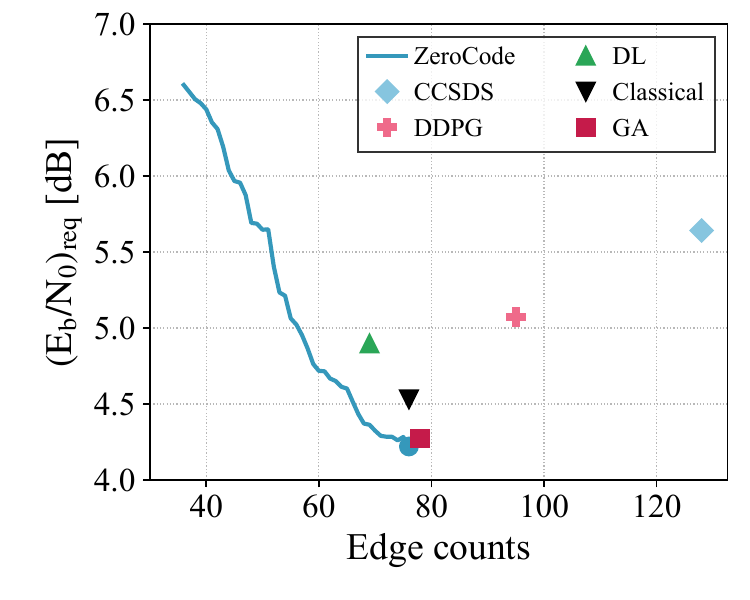}
    \caption{}
    \label{fig:library}
  \end{subfigure}\hfill
  \begin{subfigure}[b]{0.33\textwidth}
    \centering
    \includegraphics[width=\linewidth]{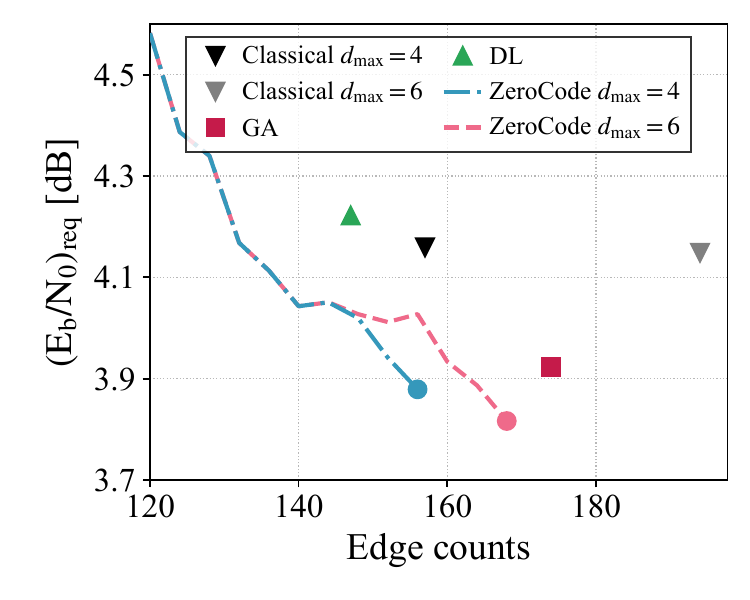}
    \caption{}
    \label{fig:maxdegree}
  \end{subfigure}\hfill
  \begin{subfigure}[b]{0.33\textwidth}
    \centering
    \includegraphics[width=\linewidth]{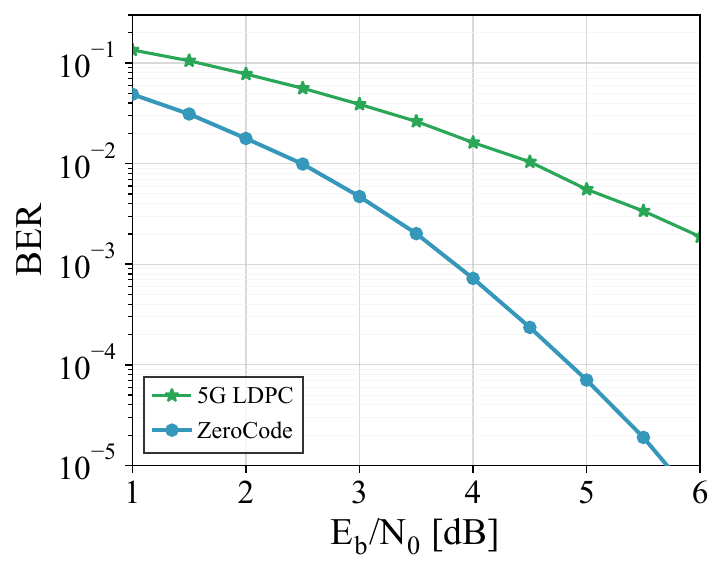}
    \caption{}
    \label{fig:5G_results}
  \end{subfigure}
  \caption{(a) Performance--complexity trade-off achieved by a single ZeroCode policy on the $(32, 16)$ code. 
  (b) The same trade-off on the $(64, 32)$ code, obtained by applying degree masks $d_{\max} \in \{4, 6\}$ at inference time to the trained policy.
  (c) BER comparison between the 5G LDPC code and ZeroCode designed under the raptor-like rate-compatible structural constraint.
  In (a)--(b), the y-axis represents the required ${\rm E}_{\rm b}/{\rm N}_0$ to achieve $\mathrm{BER} = 10^{-3}$.}
  \label{fig:mask_and_library}
\end{figure}


The max-degree masking introduced in Section~\ref{sec:masking} can also be applied at inference time without retraining. Combined with the code-library construction described in Section~\ref{sec:library}, applying a desired $d_{\max}$ constraint during rollout produces a library of degree-constrained codes spanning a wide range of edge counts.

Figure~\ref{fig:maxdegree} illustrates this on the $(64, 32)$ code. Starting from a single policy trained with the QC structural mask using lifting factor $z=4$, but without any maximum-degree constraint, we additionally apply degree masks with $d_{\max} \in \{4, 6\}$ at inference time and plot 
$({\rm E}_{\rm b}/{\rm N}_0)_{\rm req}$ as a function of the edge count. The rollout with $d_{\max} = 6$ extends further than that with the stricter constraint $d_{\max} = 4$, achieving lower $({\rm E}_{\rm b}/{\rm N}_0)_{\rm req}$ at the cost of higher edge counts. For $d_{\max}\in\{4,6\}$, ZeroCode achieves a better performance--complexity trade-off than the conventional approaches---GA~\citep{elkelesh2019decoder}, DL~\citep{choukroun2024factor}, and classical~\citep{richardson2001design}.

\subsection{5G LDPC structural constraint}
\label{sec:5G}
We further demonstrate that ZeroCode can accommodate industry-standard structural constraints through masking. As a representative case, we consider the 5G LDPC code based on BG2~\citep{3gpp38212}, whose base matrix $\mathbf{B}_{\mathrm{5G}}$ follows the raptor-like structure for rate-compatibility
\begin{equation}
\mathbf{B}_{\mathrm{5G}}=
\begin{bmatrix}
\mathbf{A} & \mathbf{B} & \mathbf{0}\\
\mathbf{C} & \mathbf{D} & \mathbf{I}
\end{bmatrix},
\label{eq:bg2_structure}
\end{equation}
where $\mathbf{B}$ is a $4 \times 4$ dual-diagonal block on the core parity columns, $\mathbf{0}$ is the $4 \times 38$ zero block, and $\mathbf{I}$ is the $38 \times 38$ identity block on the extension parity columns, all of which are fixed by the standard. The remaining blocks $(\mathbf{A}, \mathbf{C}, \mathbf{D})$ constitute the design region in which the edge pattern can be freely chosen.

ZeroCode incorporates these structural constraints through action masking. The positions in $\mathbf{B}$, $\mathbf{0}$, and $\mathbf{I}$ are permanently masked at every step. Consequently, the valid action set $\mathcal{A}_{\mathrm{valid}}(s_t)$ is restricted to the BG2 design region defined above, and QC masking is applied as well.

Figure~\ref{fig:5G_results} compares ZeroCode with the $(100,20)$ 5G LDPC code~\citep{3gpp38212} with $z=2$ on the AWGN channel. While preserving both the raptor-like rate-compatible structure and the QC structure of the 5G LDPC code, ZeroCode achieves a gain of more than $2$~dB over the 5G LDPC code at $\mathrm{BER}=10^{-3}$. This demonstrates that industry-standard structural constraints can be enforced through masking while still allowing high-performing codes to be discovered within the constrained design space.

\section{Conclusion}

In error-correcting codes, the PCM defines the code, plays a central role in decoding performance, and must often accommodate structural requirements imposed by real-world applications. To address these diverse requirements, we propose ZeroCode, an RL-based framework for flexible code design. ZeroCode formulates PCM construction as a sequential decision-making problem, in which an agent progressively builds a PCM from the all-zero matrix through valid construction actions. Experimental results show that ZeroCode substantially outperforms the prior RL-based approach and achieves superior BER performance over non-RL methods across short-to-medium block lengths.

A key component of ZeroCode is action masking, which enables structural requirements to be enforced directly throughout the construction process. Practically relevant constraints, including 4-cycle-free, quasi-cyclic, maximum-degree, and raptor-like structures, can be incorporated simply by modifying the mask without changing the underlying training procedure. Moreover, a single trained policy naturally generates a library of codes with different edge counts along a single rollout, providing a range of performance--complexity trade-offs without additional training. 

In a nutshell, ZeroCode opens a new avenue for on-demand error-correcting code construction by \emph{growing codes from zeros, guiding actions through learning, and enforcing constraints.}

\raggedbottom

\bibliographystyle{iclr2027_conference}
\bibliography{Reference}

@article{richardson2018design,
  author  = {Richardson, T. and Kudekar, S.},
  title = {Design of low-density parity-check codes for {5G New Radio}},
  journal = {IEEE Communications Magazine},
  volume  = {56},
  number  = {3},
  pages   = {28--34},
  month   = {March},
  year    = {2018},
  doi     = {10.1109/MCOM.2018.1700839}
}

@article{shor1995scheme,
  author  = {Shor, P. W.},
  title   = {Scheme for reducing decoherence in quantum computer memory},
  journal = {Physical Review A},
  volume  = {52},
  number  = {4},
  pages   = {R2493--R2496},
  month   = {October},
  year    = {1995},
  doi     = {10.1103/PhysRevA.52.R2493}
}

@article{gallager1962low,
  author  = {Gallager, R. G.},
  title   = {Low-density parity-check codes},
  journal = {IRE Transactions on Information Theory},
  volume  = {8},
  number  = {1},
  pages   = {21--28},
  month   = {January},
  year    = {1962},
  doi     = {10.1109/TIT.1962.1057683}
}

@article{mackay1999good,
  author  = {MacKay, D. J. C.},
  title   = {Good error-correcting codes based on very sparse matrices},
  journal = {IEEE Transactions on Information Theory},
  volume  = {45},
  number  = {2},
  pages   = {399--431},
  month   = {March},
  year    = {1999},
  doi     = {10.1109/18.748992}
}

@article{richardson2001design,
  author  = {Richardson, T. and Shokrollahi, M. A. and Urbanke, R.},
  title   = {Design of capacity-approaching irregular low-density parity-check codes},
  journal = {IEEE Transactions on Information Theory},
  volume  = {47},
  number  = {2},
  pages   = {619--637},
  month   = {February},
  year    = {2001},
  doi     = {10.1109/18.910578}
}

@article{elkelesh2019decoder,
  author  = {Elkelesh, A. and Ebada, M. and Cammerer, S. and Schmalen, L. and ten Brink, S.},
  title   = {Decoder-in-the-loop: Genetic optimization-based {LDPC} code design},
  journal = {IEEE Access},
  volume  = {7},
  pages   = {141161--141170},
  year    = {2019},
  doi     = {10.1109/ACCESS.2019.2942999}
}

@article{choukroun2024factor,
  author  = {Choukroun, Y. and Wolf, L.},
  title   = {Factor graph optimization of error-correcting codes for belief propagation decoding},
  journal = {arXiv preprint arXiv:2406.12900},
  year    = {2024}
}

@article{tian2025gnn,
  author  = {Tian, K. and Yue, C. and She, C. and Vucetic, B. and Li, Y.},
  title   = {{GNN}-based auto-encoder for short linear block codes: A {DRL} approach},
  journal = {IEEE Transactions on Communications},
  volume  = {73},
  number  = {10},
  pages   = {8558--8573},
  month   = {October},
  year    = {2025},
  doi     = {10.1109/TCOMM.2025.3565602}
}

@article{yue2023efficient,
  author  = {Yue, C. and Miloslavskaya, V. and Shirvanimoghaddam, M. and Vucetic, B. and Li, Y.},
  title   = {Efficient decoders for short block length codes in {6G URLLC}},
  journal = {IEEE Communications Magazine},
  volume  = {61},
  number  = {4},
  pages   = {84--90},
  month   = {April},
  year    = {2023},
  doi     = {10.1109/MCOM.001.2200275}
}

@article{shirvanimoghaddam2018short,
  author  = {Shirvanimoghaddam, M. and Mohammadi, M. S. and Abbas, R. and Minja, A. and Yue, C. and Matuz, B. and Han, G. and Lin, Z. and Liu, W. and Li, Y. and Johnson, S. J. and Vucetic, B.},
  title   = {Short block-length codes for ultra-reliable low latency communications},
  journal = {IEEE Communications Magazine},
  volume  = {57},
  number  = {2},
  pages   = {130--137},
  month   = {February},
  year    = {2019},
  doi     = {10.1109/MCOM.2018.1800181}
}

@inproceedings{lillicrap2015continuous,
  author    = {Lillicrap, T. P. and Hunt, J. J. and Pritzel, A. and Heess, N. and Erez, T. and Tassa, Y. and Silver, D. and Wierstra, D.},
  title     = {Continuous control with deep reinforcement learning},
  booktitle = {International Conference on Learning Representations},
  year      = {2016}
}

@article{schulman2017proximal,
  author  = {Schulman, J. and Wolski, F. and Dhariwal, P. and Radford, A. and Klimov, O.},
  title   = {Proximal policy optimization algorithms},
  journal = {arXiv preprint arXiv:1707.06347},
  year    = {2017}
}

@article{nachmani2018deep,
  author  = {Nachmani, E. and Marciano, E. and Lugosch, L. and Gross, W. J. and Burshtein, D. and Be'ery, Y.},
  title   = {Deep learning methods for improved decoding of linear codes},
  journal = {IEEE Journal of Selected Topics in Signal Processing},
  volume  = {12},
  number  = {1},
  pages   = {119--131},
  month   = {February},
  year    = {2018},
  doi     = {10.1109/JSTSP.2017.2788405}
}

@inproceedings{choukroun2022error,
  author    = {Choukroun, Y. and Wolf, L.},
  title     = {Error correction code transformer},
  booktitle = {Advances in Neural Information Processing Systems (NeurIPS)},
  volume    = {35},
  pages     = {38695--38705},
  year      = {2022}
}

@inproceedings{choukroun2024foundation,
  author    = {Choukroun, Y. and Wolf, L.},
  title     = {A foundation model for error correction codes},
  booktitle = {International Conference on Learning Representations (ICLR)},
  year      = {2024}
}

@inproceedings{park2024crossmpt,
  author    = {Park, S.-J. and Kwak, H.-Y. and Kim, S.-H. and Kim, Y. and No, J.-S.},
  title     = {{CrossMPT}: Cross-attention message-passing transformer for error correcting codes},
  booktitle = {International Conference on Learning Representations (ICLR)},
  year      = {2025}
}

@article{park2025multiple,
  author  = {Park, S.-J. and Kwak, H.-Y. and Kim, S.-H. and Kim, S. and Kim, Y. and No, J.-S.},
  title   = {Multiple-masks error correction code transformer for short block codes},
  journal = {IEEE Journal on Selected Areas in Communications},
  volume  = {43},
  number  = {7},
  pages   = {2518--2529},
  year    = {2025},
  doi     = {10.1109/JSAC.2025.3559154}
}

@inproceedings{habib2020learning,
  author    = {Habib, S. and Beemer, A. and Kliewer, J.},
  title     = {Learning to decode: Reinforcement learning for decoding of sparse graph-based channel codes},
  booktitle = {Advances in Neural Information Processing Systems (NeurIPS)},
  volume    = {33},
  pages     = {22396--22406},
  year      = {2020}
}

@article{habib2023reldec,
  author  = {Habib, S. and Beemer, A. and Kliewer, J.},
  title   = {{RELDEC}: Reinforcement learning-based decoding of moderate length {LDPC} codes},
  journal = {IEEE Transactions on Communications},
  volume  = {71},
  number  = {10},
  pages   = {5661--5674},
  month   = {October},
  year    = {2023},
  doi     = {10.1109/TCOMM.2023.3296621}
}

@article{berner2019dota,
  author  = {{OpenAI} and Berner, C. and Brockman, G. and Chan, B. and
             Cheung, V. and D{\k{e}}biak, P. and Dennison, C. and
             Farhi, D. and Fischer, Q. and Hashme, S. and Hesse, C. and
             J{\'o}zefowicz, R. and Gray, S. and Olsson, C. and
             Pachocki, J. and Petrov, M. and Pinto, H. P. d. O. and
             Raiman, J. and Salimans, T. and Schlatter, J. and
             Schneider, J. and Sidor, S. and Sutskever, I. and
             Tang, J. and Wolski, F. and Zhang, S.},
  title   = {Dota 2 with Large Scale Deep Reinforcement Learning},
  journal = {arXiv preprint arXiv:1912.06680},
  year    = {2019}
}

@article{mirhoseini2021graph,
  author  = {Mirhoseini, A. and Goldie, A. and Yazgan, M. and Jiang, J. W. and Songhori, E. and Wang, S. and Lee, Y.-J. and Johnson, E. and Pathak, O. and Nova, A. and Pak, J. and Tong, A. and Srinivasa, K. and Hang, W. and Tuncer, E. and Le, Q. V. and Laudon, J. and Ho, R. and Carpenter, R. and Dean, J.},
  title   = {A graph placement methodology for fast chip design},
  journal = {Nature},
  volume  = {594},
  number  = {7862},
  pages   = {207--212},
  month   = {June},
  year    = {2021},
  doi     = {10.1038/s41586-021-03544-w}
}

@inproceedings{huang2020closer,
  author    = {Huang, S. and Onta{\~n}{\'o}n, S.},
  title     = {A closer look at invalid action masking in policy gradient algorithms},
  booktitle = {Proceedings of the Thirty-Fifth International Florida Artificial Intelligence Research Society Conference (FLAIRS)},
  volume    = {35},
  year      = {2022},
  doi       = {10.32473/flairs.v35i.130584}
}

@article{fan2008design,
  author  = {Fan, J. and Xiao, Y. and Kim, K.},
  title   = {Design {LDPC} codes without cycles of length 4 and 6},
  journal = {Research Letters in Communications},
  volume  = {2008},
  pages   = {354137},
  year    = {2008},
  doi     = {10.1155/2008/354137}
}

@techreport{ccsds2015short,
  author      = {{Consultative Committee for Space Data Systems (CCSDS)}},
  title       = {Short block length {LDPC} codes for {TC} synchronization and channel coding},
  institution = {CCSDS},
  number      = {231.1-O-1},
  type        = {Orange Book},
  month       = {April},
  year        = {2015}
}

@article{hu2005regular,
  author  = {Hu, X. Y. and Eleftheriou, E. and Arnold, D. M.},
  title   = {Regular and irregular progressive edge-growth {Tanner} graphs},
  journal = {IEEE Transactions on Information Theory},
  volume  = {51},
  number  = {1},
  pages   = {386--398},
  month   = {January},
  year    = {2005},
  doi     = {10.1109/TIT.2004.839541}
}

@inproceedings{kwak2023boosting,
  author    = {Kwak, H.-Y. and Yun, D.-Y. and Kim, Y. and Kim, S.-H. and No, J.-S.},
  title     = {Boosting learning for {LDPC} codes to improve the error-floor performance},
  booktitle = {Advances in Neural Information Processing Systems (NeurIPS)},
  volume    = {36},
  pages     = {22115--22131},
  year      = {2023}
}

@book{ryan2009channel,
  author    = {Ryan, W. E. and Lin, S.},
  title     = {Channel Codes: Classical and Modern},
  publisher = {Cambridge University Press},
  year      = {2009},
  isbn      = {978-0-521-84868-8},
  doi       = {10.1017/CBO9780511803253}
}

@inproceedings{chen2019large,
  author    = {Chen, H. and Dai, X. and Cai, H. and Zhang, W. and Wang, X. and Tang, R. and Zhang, Y. and Yu, Y.},
  title     = {Large-scale interactive recommendation with tree-structured policy gradient},
  booktitle = {Proceedings of the AAAI Conference on Artificial Intelligence},
  volume    = {33},
  pages     = {3312--3320},
  year      = {2019},
  doi       = {10.1609/aaai.v33i01.33013312}
}

@inproceedings{zhao2013ldpc,
  author    = {Zhao, K. and Zhao, W. and Sun, H. and Zhang, X. and Zheng, N. and Zhang, T.},
  title     = {{LDPC-in-SSD}: Making advanced error correction codes work effectively in solid state drives},
  booktitle = {11th USENIX Conference on File and Storage Technologies (FAST 13)},
  pages     = {243--256},
  year      = {2013}
}

@article{ramkumar2022codes,
  author  = {Ramkumar, V. and Balaji, S. B. and Sasidharan, B. and Vajha, M. and Krishnan, M. N. and Kumar, P. V.},
  title   = {Codes for distributed storage},
  journal = {Foundations and Trends in Communications and Information Theory},
  volume  = {19},
  number  = {4},
  pages   = {547--813},
  year    = {2022},
  doi     = {10.1561/0100000115}
}

@techreport{3gpp38212,
  author      = {{3GPP}},
  title       = {{NR}; Multiplexing and channel coding},
  type        = {Technical Specification},
  institution = {3GPP},
  number      = {TS 38.212, Version 15.6.0, Release 15},
  month       = {July},
  year        = {2019}
}

@article{trifonov2012efficient,
  author  = {Trifonov, P.},
  title   = {Efficient Design and Decoding of Polar Codes},
  journal = {IEEE Transactions on Communications},
  volume  = {60},
  number  = {11},
  pages   = {3221--3227},
  month   = {November},
  year    = {2012},
  doi     = {10.1109/TCOMM.2012.081512.110872}
}

@techreport{3gpp2016polar,
  author      = {{Huawei Technologies Co., Ltd.} and {HiSilicon Technologies Co., Ltd.}},
  title       = {Polar Code Design and Rate Matching},
  institution = {3rd Generation Partnership Project (3GPP)},
  number      = {R1-167209},
  address     = {Gothenburg, Sweden},
  month       = {August},
  year        = {2016},
  note        = {Meeting \#86}
}

@inproceedings{he2017beta,
  author    = {He, G. and Belfiore, J.-C. and Land, I. and Yang, G. and Liu, X. and Chen, Y. and Li, R. and Wang, J. and Ge, Y. and Zhang, R. and Tong, W.},
  title = {{$\beta$-Expansion}: A theoretical framework for fast and recursive construction of polar codes},
  booktitle = {IEEE Global Communications Conference (GLOBECOM)},
  pages     = {1--6},
  month     = {December},
  year      = {2017},
  doi       = {10.1109/GLOCOM.2017.8254146}
}

@article{huang2020ai,
  author  = {Huang, L. and Zhang, H. and Li, R. and Ge, Y. and Wang, J.},
  title   = {{AI} Coding: Learning to Construct Error Correction Codes},
  journal = {IEEE Transactions on Communications},
  volume  = {68},
  number  = {1},
  pages   = {26--39},
  month   = {January},
  year    = {2020},
  doi     = {10.1109/TCOMM.2019.2951403}
}

@inproceedings{schulman2015high,
  author    = {Schulman, John and Moritz, Philipp and Levine, Sergey and Jordan, Michael and Abbeel, Pieter},
  title     = {High-Dimensional Continuous Control Using Generalized Advantage Estimation},
  booktitle = {International Conference on Learning Representations (ICLR)},
  year      = {2016}
}

@article{stable-baselines3,
  author  = {Raffin, Antonin and Hill, Ashley and Gleave, Adam and Kanervisto, Anssi and Ernestus, Maximilian and Dormann, Noah},
  title = {{Stable-Baselines3}: Reliable Reinforcement Learning Implementations},
  journal = {Journal of Machine Learning Research},
  volume  = {22},
  number  = {268},
  pages   = {1--8},
  year    = {2021}
}

@article{silver2016mastering,
  author  = {Silver, David and Huang, Aja and Maddison, Chris J. and Guez, Arthur and Sifre, Laurent and van den Driessche, George and Schrittwieser, Julian and Antonoglou, Ioannis and Panneershelvam, Veda and Lanctot, Marc and Dieleman, Sander and Grewe, Dominik and Nham, John and Kalchbrenner, Nal and Sutskever, Ilya and Lillicrap, Timothy and Leach, Madeleine and Kavukcuoglu, Koray and Graepel, Thore and Hassabis, Demis},
  title = {Mastering the game of {Go} with deep neural networks and tree search},
  journal = {Nature},
  volume  = {529},
  number  = {7587},
  pages   = {484--489},
  year    = {2016},
  doi     = {10.1038/nature16961}
}

@article{kwak2025boosted,
  author  = {Kwak, Hee-Youl and Yun, Dae-Young and Kim, Yongjune and Kim, Sang-Hyo and No, Jong-Seon},
  title   = {Boosted Neural Decoders: Achieving Extreme Reliability of {LDPC} Codes for {6G} Networks},
  journal = {IEEE Journal on Selected Areas in Communications},
  volume  = {43},
  number  = {4},
  pages   = {1089--1102},
  month   = {April},
  year    = {2025},
  doi     = {10.1109/JSAC.2025.3531553}
}
\appendix

\section{Training of ZeroCode}
\label{app:training_details}

\subsection{Training procedure}

\begin{algorithm}[H]
\caption{Training procedure of ZeroCode}
\label{alg:ppo_full}
\DontPrintSemicolon
\KwIn{PCM dimensions $(m,n)$, structural constraints, rollout size $N$, mini-batch size $b$, epochs $K$, total timesteps $T$}
\KwOut{Trained policy $\pi_\theta$}

Initialize $\theta$ and $\phi$; set $\mathcal{D}\leftarrow\emptyset$, $\mathbf{H}_t\leftarrow\mathbf{0}_{m\times n}$, and global step $\leftarrow 0$\;

\While{global step $< T$}{
  \tcc{Rollout collection}
  \While{$|\mathcal{D}|<N$}{
    Construct the action mask $\mathbf{M}(\mathbf{H}_t)$ to enforce the given structural constraints, as described in Section~\ref{sec:masking}\;
    Sample $a_t \sim \tilde{\pi}_\theta(\cdot\mid\mathbf{H}_t)$ from the masked policy in Eq.~\eqref{eq:masked_policy}\;
    Apply the sampled action to $\mathbf{H}_t$ to obtain $\mathbf{H}_{t+1}$\;
    compute $R_t=\Lambda(\mathbf{H}_{t+1})$ as in Eq.~\eqref{eq:reward}\;
    Set $\mathrm{done}_t$ to true when $s_{t+1}$ has no valid actions, and false otherwise\;
    Store $(\mathbf{H}_t,a_t,R_t,\mathbf{H}_{t+1},\mathrm{done}_t)$ in $\mathcal{D}$\;  
    global step $\mathrel{+}=1$\;
    \uIf{$\mathrm{done}_t$}{
        $\mathbf{H}_t\leftarrow\mathbf{0}_{m\times n}$\;
    }
    \Else{
        $\mathbf{H}_t\leftarrow\mathbf{H}_{t+1}$\;
    }
  }

  \tcc{Policy and value update}
  Compute GAE advantages $\{\hat{A}_t\}$ and target returns from $\mathcal{D}$\;
  Update $(\theta,\phi)$ for $K$ epochs using mini-batches of size $b$ according to Eq.~\eqref{eq:ppo_loss}\;
  $\mathcal{D}\leftarrow\emptyset$\;
}
\Return $\pi_\theta$\;
\end{algorithm}

Algorithm~\ref{alg:ppo_full} summarizes the complete training procedure of ZeroCode. Each episode starts from the all-zero PCM $\mathbf{H}_0=\mathbf{0}_{m\times n}$. At step $t$, the action mask $\mathbf{M}(\mathbf{H}_t)$ is constructed, and one valid edge is sampled from $\tilde{\pi}_\theta(\cdot\mid\mathbf{H}_t)$ and added to the current PCM. The resulting PCM $\mathbf{H}_{t+1}$ is then evaluated by the BP decoder. The resulting reward and transition are appended to the rollout buffer. The episode terminates when no valid edge remains in the masked action set. The environment is then reset.

Rollout collection continues across episodes until the buffer contains $N$ transitions. We then compute target returns and generalized-advantage estimates and update the policy and value networks for $K$ optimization epochs using mini-batches under the PPO objective. The buffer is cleared after each update. This process is repeated until the total number of training timesteps reaches $T$, yielding the trained policy $\pi_\theta$.

\subsection{Optimization details}
\begin{table}[h]
\centering
\caption{Hyperparameters used to train ZeroCode.}
\label{tab:hyperparams}
\renewcommand{\arraystretch}{1.08}
\begin{tabular}{@{\hspace{10pt}}l@{\hspace{24pt}}r@{\hspace{10pt}}}
\toprule
\textbf{Parameter} & \textbf{Value} \\
\midrule
Learning rate & $10^{-5}$ \\
Total timesteps $T$ & $3\times10^{6}$ \\
Rollout size $N$ & $2{,}048$ \\
Number of epochs $K$ & $10$ \\
Mini-batch size $b$ & $64$ \\
Discount factor $\gamma$ & $0.99$ \\
GAE parameter $\lambda$ & $0.95$ \\
Clip range $\epsilon$ & $0.2$ \\
Value loss coefficient $c_1$ & $0.5$ \\
Entropy coefficient $c_2$ & $0.01$ \\
Max gradient norm & $0.5$ \\
Target KL divergence & $0.02$ \\
Policy hidden layers $\pi_\theta$ & $[4096,\ 2048]$ \\
Value hidden layers $V_\phi$ & $[2048,\ 1024]$ \\
\bottomrule
\end{tabular}
\end{table}

Table~\ref{tab:hyperparams} reports the optimization settings used in all experiments. For each PPO update, we collect $2{,}048$ transitions and perform $10$ optimization epochs with mini-batches of $64$. We use a learning rate of $10^{-5}$ for both the policy and value networks. Advantages are estimated with GAE~\citep{schulman2015high}, using discount factor $\gamma=0.99$ and $\lambda=0.95$.

The training objective combines the clipped policy loss, the value-function loss, and an entropy bonus:

\begin{equation}
\label{eq:ppo_loss}
L_{\mathrm{PPO}}(\theta,\phi)
=
\hat{\mathbb{E}}_t
\left[
-
L_t^{\mathrm{CLIP}}(\theta)
+
c_1 L_t^{\mathrm{VF}}(\phi)
-
c_2 S[\tilde{\pi}_\theta](s_t)
\right].
\end{equation}

We set the clipping parameter to $\epsilon=0.2$, the value-loss coefficient to $c_1=0.5$, and the entropy coefficient to $c_2=0.01$. For optimization stability, the gradient norm is clipped at $0.5$. We also use a target KL divergence of $0.02$ and terminate the remaining epochs of an update when this threshold is exceeded.


\paragraph{Networks and compute.}
The policy $\pi_\theta$ and value function $V_\phi$ are represented by separate two-layer MLPs. Their hidden-layer sizes are $[4096, 2048]$ and $[2048, 1024]$, respectively. We use the MaskablePPO implementation in Stable-Baselines3 Contrib v2.7.1~\citep{stable-baselines3}, which supports invalid-action masking. All experiments run on one NVIDIA RTX A5000 GPU with 24 GB of memory; depending on the code configuration, training one policy takes approximately 4--18 hours.

\subsection{Training cost}
\label{app:training_cost}

\begin{table}[h]
\centering
\caption{Approximate wall-clock training cost for the $(32,16)$ code. The code library contains the $L=41$ valid PCMs from the first valid design at 36 edges to the best design at 76 edges. For GA and DL, library costs are estimated as $L$ times the single-PCM cost, assuming independent optimization for each target edge count.}
\label{tab:training_cost}
\begin{tabular}{lcc}
\toprule
Method & Cost for a single PCM & Cost for a library ($L=41$) \\
\midrule
ZeroCode & 4 hours ($t_a$) & 4 hours ($t_a$) \\
GA~\citep{elkelesh2019decoder} & 6 hours ($t_b$) & 246 hours ($L t_b$) \\
DL~\citep{choukroun2024factor} & 2 hours ($t_c$) & 82 hours ($L t_c$) \\
\bottomrule
\end{tabular}
\end{table}

For the $(32,16)$ code, the GA runtime was measured using its public implementation and is comparable to that of ZeroCode because both methods repeatedly evaluate candidate PCMs with a BP decoder. The DL method has a shorter optimization time, but yielded lower BER performance in our experiments.

For a single target PCM, ZeroCode has a training cost comparable
to GA. Under an independent optimization protocol for each target
edge count, the estimated costs of constructing an $L=41$ code
library are $L t_b$ for GA and $L t_c$ for DL.
ZeroCode instead obtains all $L=41$ valid intermediate PCMs
along a rollout of a single trained policy, providing multiple
performance--complexity operating points without retraining.

\subsection{Detailed Experimental Settings and Statistical Stability}
\label{app:seed_robustness}

\begin{figure}[t]
  \centering
  \includegraphics[width=0.42\linewidth]{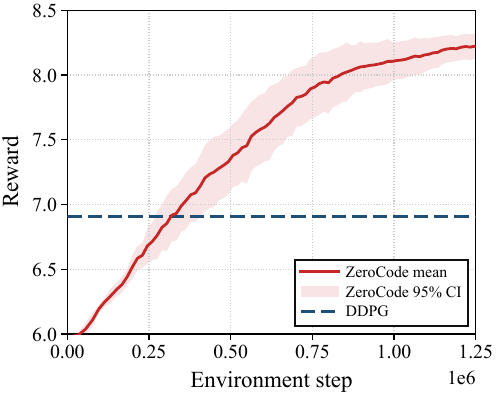}
  \caption{Training robustness of ZeroCode across eight random seeds for the
  $(32,16)$ code. The solid red curve shows the mean best-PCM reward
  $R_{\mathrm{best}}=-\ln(\mathrm{BER}_{\mathrm{best}})$ across seeds. The
  shaded region denotes the 95\% confidence interval across eight random seeds.
  Rewards are evaluated at
  ${\rm E}_{\rm b}/{\rm N}_0=5.0$~dB and averaged over non-overlapping bins of 200
  episodes. The horizontal dashed line denotes the reward of the final PCM
  produced by DDPG~\citep{tian2025gnn} under the same evaluation setting.}
  \label{fig:seed_robustness}
\end{figure}

\paragraph{MLP architecture.}
The MLP implementation follows the default implementation choices of Stable-Baselines3 Contrib. Specifically, Tanh is used as the activation function for the hidden layers, Adam is used as the optimizer, and the network weights are initialized using orthogonal initialization.

\paragraph{GNN architecture.}
For the GNN-based architectures evaluated in Appendix~\ref{sec:architecture_ablation}, we follow the GNN configuration of~\citet{tian2025gnn}. Each PCM element is treated as a node in a lattice graph and exchanges messages with its four neighboring elements. The embedding and message dimensions are both set to $10$. The GNN consists of $3$ message-passing layers, and the MLPs used for message computation and embedding updates consist of $3$ layers with $40$ hidden units.

\paragraph{Channel model.}
Each codeword $\mathbf{c}\in\{0,1\}^{n}$ is BPSK-modulated as $\mathbf{x}=\mathbf{1}-2\mathbf{c}$ and transmitted over an AWGN channel, yielding $\mathbf{y}=\mathbf{x}+\mathbf{z}$, where $\mathbf{z}\sim\mathcal{N}(\mathbf{0},\sigma^{2}\mathbf{I})$ and $\sigma^{2}=1/\left(2r\,{\rm E}_{\rm b}/{\rm N}_0\right)$. The resulting channel LLRs, $\mathbf{m}^{\mathrm{ch}}=2\mathbf{y}/\sigma^{2}$, are provided as input to the BP decoder. The polar-code experiment in Figure~\ref{fig:generalization}(b) instead uses QPSK modulation, with channel conditions specified in terms of ${\rm E}_{\rm s}/{\rm N}_0$.

\paragraph{Reward and evaluation.}
Since ZeroCode sequentially constructs the PCM from the all-zero matrix, the decoding performance of intermediate PCMs is also used as the reward signal during training. For all PCMs encountered along the construction trajectory, we fix the code rate to the target rate $r=k/n$, regardless of the current rank of $\mathbf{H}_t$. During training, the error rate used to compute the reward is estimated by continuing the simulation until $100$ frame errors are observed. For the final performance curves, the simulation at each evaluated SNR point is continued until $1{,}000$ frame errors are observed.

\paragraph{Baseline implementation.}
For DDPG~\citep{tian2025gnn}, we directly use the PCMs provided by the publicly available implementation. For GA~\citep{elkelesh2019decoder}, we use the publicly available implementation and follow its original optimization procedure. Specifically, the $20$ best-performing PCMs in each population are retained as parents for mutation and crossover, and the optimization is performed for $500$ generations. For DL~\citep{choukroun2024factor}, we perform $20$ optimization steps, with the line search at each step restricted to the $110$ smallest candidate step sizes.

\paragraph{Statistical stability.}
To assess sensitivity to random initialization, we independently train ZeroCode with eight random seeds for the $(32,16)$ code. As shown in Figure~\ref{fig:seed_robustness}, the ZeroCode confidence interval rises above the DDPG reference after the initial exploration phase, and all eight runs end above this reference. These results indicate stable training behavior across random seeds.

\begin{figure}[t]
  \centering
  \begin{subfigure}[b]{0.41\textwidth}
    \centering
    \includegraphics[width=\linewidth]{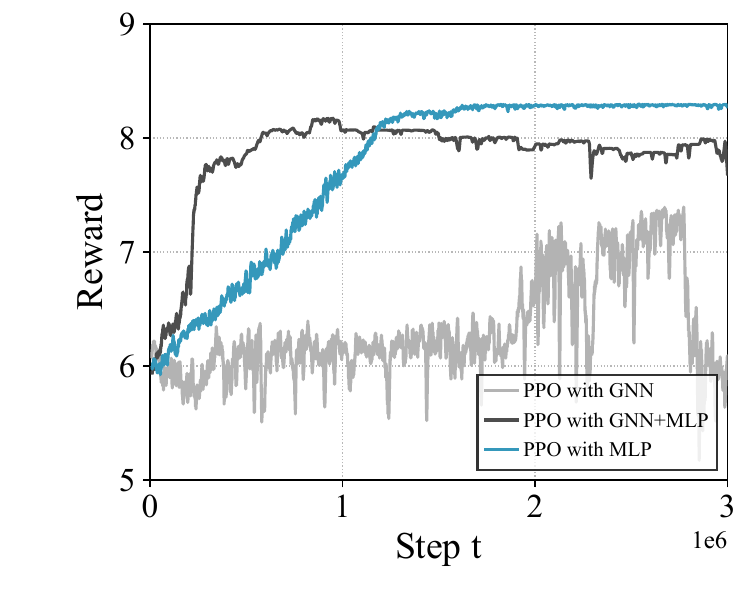}
    \caption{}
    \label{fig:training_curve}
  \end{subfigure}
  \begin{subfigure}[b]{0.42\textwidth}
    \centering
    \includegraphics[width=\linewidth]{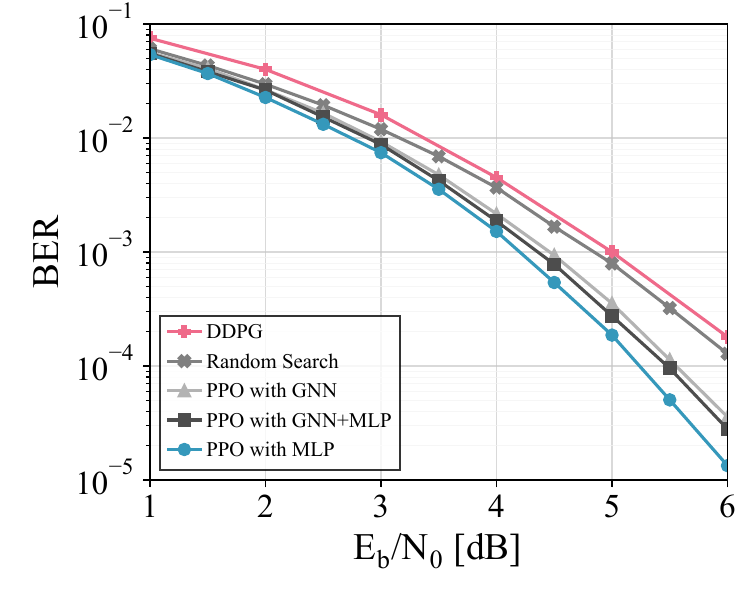}
    \caption{}
    \label{fig:plot_embedding}
  \end{subfigure}  
    \caption{Comparison of three actor--critic architectures (GNN, GNN+MLP, and MLP) on the $(32, 16)$ code. (a) Training curves show the 100-episode moving average of the reward $R_{\mathrm{best}}=-\ln(\mathrm{BER}_{\mathrm{best}})$ of the best PCM encountered in each episode. (b) Performance of the resulting codes constructed by each architecture.}
  \label{fig:embedding_comparison}
\end{figure}

\section{Actor--critic architecture}
\label{sec:architecture_ablation}

The most closely related RL-based method, \citet{tian2025gnn}, employs a GNN-based actor--critic architecture. Since PCM construction is naturally associated with graph structures, incorporating a graph-based representation may appear well motivated. However, it remains unclear whether an explicit graph embedding is necessary for the sequential PCM construction problem considered in ZeroCode. To investigate this, we compare three actor--critic architectures with different state representations: (i) \textbf{GNN}, in which a GNN directly performs both state embedding and policy/value estimation; (ii) \textbf{GNN+MLP}, in which a GNN first embeds the current state and the resulting embedding is processed by an MLP-based actor--critic; and (iii) \textbf{MLP}, in which the current PCM is flattened and fed directly into an MLP without any learnable graph embedding as shown in Section~\ref{sec:policy_optimization}.

Figure~\ref{fig:training_curve} compares the training curves of the three architectures on the $(32,16)$ code using the best-PCM reward $R_{\mathrm{best}}$. The GNN model remains at a low reward throughout training, suggesting that the GNN representation may be less effective for the fine-grained action selection required in this setting. The GNN+MLP model achieves faster reward growth during the early stage of training, indicating that graph-based embeddings help stabilize early exploration; however, its performance saturates relatively early. In contrast, the MLP model improves more steadily and ultimately achieves the highest reward. The same trend is observed in Figure~\ref{fig:plot_embedding}, where the MLP-based architecture achieves the best BER performance among the three.

These results suggest that an explicit graph embedding is not necessary for ZeroCode. Since the current PCM $\mathbf{H}_t$ already preserves the positional information of candidate edges, the MLP-based actor--critic can learn the relevant structural patterns directly from the flattened input. Accordingly, ZeroCode adopts the MLP-based architecture.

\section{Action masking ablation}
\label{sec:action_masking_ablation}

\begin{table}[t]
  \centering
  \caption{Comparison of action masking strategies. BER is evaluated at ${\rm E}_{\rm b}/{\rm N}_0=5.0$ dB using a BP decoder; the lowest BER for each block length is shown in bold.}
  \label{tab:exp_setup}
  \setlength{\tabcolsep}{4pt}
  \begin{tabular}{c l c c c c c}
    \toprule
    $(n,k)$ & Action mask & Edges & $d_{\max}$ & 4-cycle & 6-cycle & BER \\
    \midrule
    \multirow{2}{*}{$(32,16)$}
      & 4-cycle
      & 76 & 5 & 0 & 88
      & $\mathbf{1.81\times10^{-4}}$ \\
      & Max-degree ($d_{\max}=8$)
      & 76 & 5 & 3 & 99
      & $2.24\times10^{-4}$ \\
    \midrule
    \multirow{3}{*}{$(64,32)$}
      & 4-cycle       & 178 & 6 & 0 & 311 & $6.76\times10^{-5}$ \\
      & Max-degree ($d_{\max}=8$) & 176 & 7 & 5 & 296 & $7.25\times10^{-5}$ \\
      & QC ($Z=8$)    & 168 & 4 & 0 & 144 & $\mathbf{4.23\times10^{-5}}$ \\
    \midrule
    \multirow{3}{*}{$(128,64)$}
      & 4-cycle       & 400 & 8 & 0  & 818 & $1.15\times10^{-5}$ \\
      & Max-degree ($d_{\max}=8$) & 394 & 6 & 14 & 572 & $1.81\times10^{-5}$ \\
      & QC ($Z=16$)   & 368 & 4 & 0  & 192 & $\mathbf{5.44\times10^{-6}}$ \\
    \midrule
    \multirow{3}{*}{$(256,128)$}
      & 4-cycle       & 864 & 7 & 0  & 870 & $2.29\times10^{-6}$ \\
      & Max-degree ($d_{\max}=8$) & 833 & 7 & 21 & 705 & $3.14\times10^{-6}$ \\
      & QC ($Z=16$)   & 848 & 5 & 0 & 720 & $\mathbf{6.49\times10^{-7}}$ \\
    \bottomrule
  \end{tabular}
\end{table}

To compare different action masking strategies, we independently train policies with 4-cycle, maximum-degree, and QC masking for rate-\(1/2\) codes with block lengths \(n \in \{32,64,128,256\}\). QC masking is considered only for \(n \geq 64\), given the small matrix size at \(n=32\). For each policy, the best-performing PCM is evaluated at ${\rm E}_{\rm b}/{\rm N}_0=5.0$~dB using BP decoding with eight iterations for \(n=32\) and five iterations for the other block lengths. Table~\ref{tab:exp_setup} summarizes the resulting BER, edge count, maximum variable-node degree, and numbers of 4- and 6-cycles.

The results confirm that the constraints imposed by each action mask are satisfied as intended: 4-cycle masking eliminates all 4-cycles, maximum-degree masking produces PCMs with \(d_{\max}\leq 8\), and QC masking preserves the prescribed quasi-cyclic structure. QC masking achieves the lowest BER for \(n \in \{64,128,256\}\), while 4-cycle masking performs best among the evaluated strategies for \(n=32\). Accordingly, the ZeroCode performance curves in Figure~\ref{fig:performance} use the best-performing masking strategy for each block length: 4-cycle masking for \(n=32\) and QC masking for the remaining block lengths.

\section{Generality}
\label{sec:generalization}

The main experiments focus on PCM construction with BP decoding over the AWGN channel. To examine whether ZeroCode extends beyond this channel and code–decoder combination, we conduct two auxiliary experiments involving a different channel model and a different code family.

\begin{figure}[H]
  \vspace{-1em}
  \centering
  \begin{subfigure}[b]{0.425\textwidth}
    \centering
    \includegraphics[width=\linewidth]{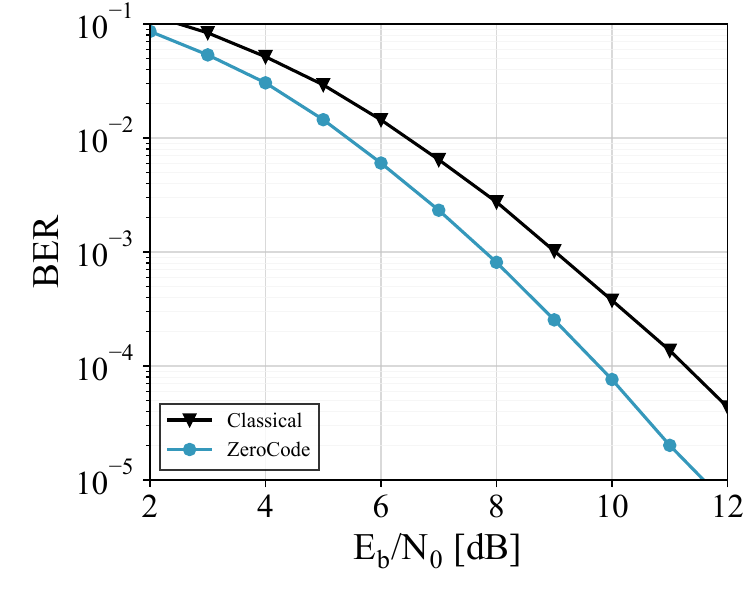}
    \caption{}
    \label{fig:rayleigh_performance}
  \end{subfigure}%
  \hspace{0.01\textwidth}%
  \begin{subfigure}[b]{0.425\textwidth}
    \centering
    \includegraphics[width=\linewidth]{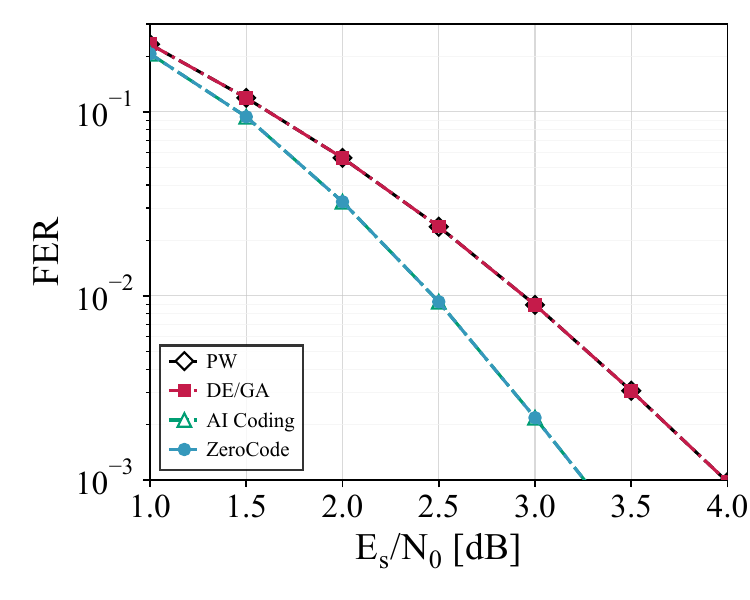}
    \caption{}
    \label{fig:polar_performance}
  \end{subfigure}
  \caption{Generality of ZeroCode:
  (a) BER comparison between ZeroCode and the classical LDPC construction over
  a Rayleigh fading channel; (b) FER comparison of the ZeroCode polar-code
  construction with AI Coding, PW, and DE/GA over the AWGN channel.}
  \label{fig:generalization}
  \vspace{-0.5em}
\end{figure}

First, we evaluate the $(32,16)$ LDPC code constructed by ZeroCode over a Rayleigh fading channel and compare it with the classical $(32,16)$ LDPC code used in the AWGN experiment. As shown in Figure~\ref{fig:generalization}(a), ZeroCode achieves a lower BER than the classical construction throughout the evaluated ${\rm E}_{\rm b}/{\rm N}_0$ range, indicating that its performance gain extends beyond the AWGN channel.

Second, we apply ZeroCode’s sequential decision-making principle to polar-code construction by treating the selection of information-bit positions as a sequence of discrete actions. ZeroCode uses PPO to learn a policy for constructing the polar information set. The state is a length-\(n\) binary vector whose ones denote the selected information-bit positions. At each step, an action selects one of the remaining frozen positions and sets the corresponding state entry to one. For a target \((n,k)\) code, the episode terminates after \(k\) positions have been selected. The policy is trained using the negative logarithm of the frame error rate (FER) measured at ${\rm E}_{\rm s}/{\rm N}_0=3.0$ dB as the reward.

We compare ZeroCode with AI Coding~\citep{huang2020ai}, polarization weight (PW)~\citep{3gpp2016polar,he2017beta}, and density evolution/Gaussian approximation (DE/GA)~\citep{trifonov2012efficient}. All methods are evaluated on a \((128,64)\) polar code with QPSK modulation over the AWGN channel using the same path-metric-based successive-cancellation list (SCL-PM) decoder with list size $8$. For this configuration, PW and DE/GA select the same information-bit set and therefore yield identical FER performance. As shown in Figure~\ref{fig:generalization}(b), ZeroCode matches the reported FER performance of AI Coding, and both outperform PW and DE/GA over the evaluated ${\rm E}_{\rm s}/{\rm N}_0$ range. Together, these auxiliary results demonstrate the applicability of ZeroCode’s sequential construction approach beyond a single channel model or code–decoder combination.

\section{Limitations}
\label{sec:Limitations}

This work primarily focuses on PCM construction with BP decoding. While Appendix~\ref{sec:generalization} demonstrates preliminary generalization to a different channel model and to polar-code construction, the broader applicability of ZeroCode has not yet been systematically evaluated. Extending and validating the framework across diverse decoders, channels, and code representations is an important direction for future work.

\end{document}